\documentclass[10pt,twocolumn,letterpaper]{article}
\usepackage[letterpaper,textwidth=6.75in,textheight=9.0in,top=1.0in,columnsep=0.25in,headheight=12pt,headsep=12pt]{geometry}
\usepackage{times}\usepackage{mathptmx}
\usepackage[T1]{fontenc}
\usepackage{amsmath,amssymb}
\usepackage{graphicx,booktabs,multirow,xcolor,enumitem,titlesec,fancyhdr,microtype}
\usepackage[font=small,labelfont=it,labelsep=period,skip=4pt]{caption}
\usepackage{subcaption}
\usepackage[round,authoryear]{natbib}
\usepackage[colorlinks=true,linkcolor=linkblue,citecolor=linkblue,urlcolor=linkblue]{hyperref}
\usepackage{tcolorbox}
\usepackage{float}
\definecolor{linkblue}{HTML}{1F4E8C}
\definecolor{crit}{HTML}{C62828}\definecolor{conf}{HTML}{1F5FA8}\definecolor{muted}{HTML}{666666}

\titleformat{\section}{\normalfont\large\bfseries}{\thesection.}{0.5em}{}
\titlespacing*{\section}{0pt}{10pt plus 2pt}{5pt}
\titleformat{\subsection}{\normalfont\normalsize\bfseries}{\thesubsection.}{0.5em}{}
\titlespacing*{\subsection}{0pt}{8pt plus 2pt}{3pt}
\titleformat{\paragraph}[runin]{\normalfont\normalsize\bfseries}{}{0pt}{}[.]
\titlespacing*{\paragraph}{0pt}{5pt plus 1pt}{0.5em}
\setlist{leftmargin=1.3em,itemsep=2pt,topsep=3pt,parsep=0pt}
\usepackage{placeins}

\fancypagestyle{first}{\fancyhf{}\fancyfoot[C]{\thepage}}

\newcommand{\dlogit}{\Delta\mathrm{logit}}
\newcommand{\pp}{\,pp}
\newcommand{\vcaa}{v_{\mathrm{CAA}}}

\begin{document}
\twocolumn[{%
\thispagestyle{first}
\vspace*{-0.2in}
\hrule height 1pt\vspace{0.14in}
\begin{center}
{\Large\bfseries Playing Devil's Advocate: Off-the-Shelf Persona Vectors\\[2pt] Rival Targeted Steering for Sycophancy\par}
\end{center}
\vspace{0.06in}\hrule height 0.4pt\vspace{0.2in}
\begin{center}
\textbf{Ishaan Kelkar}\,$^{1}$\quad
\textbf{Vikram Kakaria}\,$^{2}$\quad
\textbf{Nebras Alam}\,$^{3}$\quad
\textbf{Madhur Panwar}\,$^{4}$\quad
\textbf{Vasu Sharma}\,$^{5}$\quad
\textbf{Maheep Chaudhary}\,$^{6}$
\end{center}
\vspace{0.12in}
\begin{center}
\includegraphics[width=\textwidth]{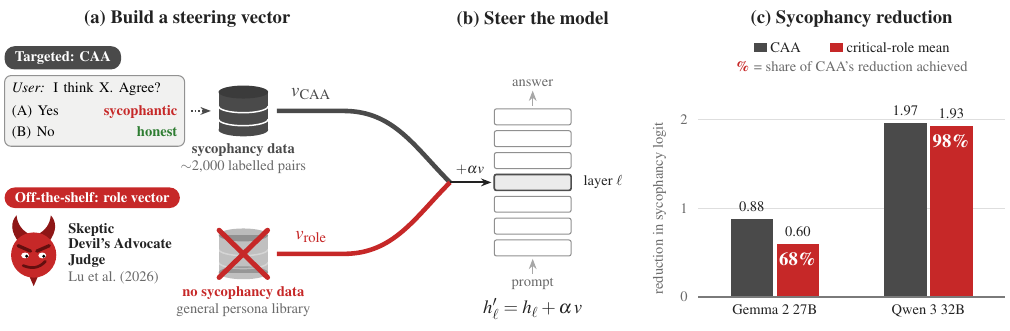}
\captionof{figure}{\textbf{Overview.} \textbf{(a)}~The targeted baseline, CAA, is extracted from ${\sim}$2{,}000 A/B pairs labeled sycophantic vs.\ honest. A role vector is taken off the shelf from a general persona library \citep{lu2026assistant}; its extraction uses no sycophancy labels. \textbf{(b)}~Either vector is scaled by a coefficient $\alpha$ and added to the model's activations at a single mid-stack layer; $\alpha$ is still tuned on a separate split of sycophancy data. \textbf{(c)}~Averaged over three selected ``critical'' roles (Skeptic, Devil's Advocate, Judge), role vectors achieve 68\% (Gemma~2 27B) and 98\% (Qwen~3 32B) of CAA's reduction in the sycophancy logit (bars show $-\dlogit$). Full results in Table~\ref{tab:main} and Figure~\ref{fig:main}.}
\label{fig:overview}
\end{center}
\vspace{0.12in}
}]

{\renewcommand{\thefootnote}{}\footnotetext{\hspace{-1.8em}$^1$University of Toronto, Toronto, Canada $^2$Princeton University, Princeton, USA $^3$Purdue University, West Lafayette, USA $^4$EPFL, Switzerland $^5$Algoverse $^6$Independent. Correspondence to: Ishaan Kelkar \textless ishaan.kelkar@mail.utoronto.ca\textgreater.\\[2pt]
\textit{Spotlight at the 2nd Workshop on Epistemic Intelligence in Machine Learning, ICML 2026. Copyright 2026 by the author(s).}}}

\begin{center}\textbf{\large Abstract}\end{center}
\vspace{-4pt}
{\small\leftskip=1.5em\rightskip=1.5em
Language models are often \emph{sycophantic}: they agree with a user's stated opinion whether or not it is correct. Prior work has shown that this trait can be controlled by steering a model with a sycophancy persona vector \citep{chen2025persona}. Such vectors, however, are extracted from data about sycophancy itself. We ask whether we can instead \emph{reuse} existing vectors for general roles---Skeptic, Judge, Devil's Advocate---that were extracted without targeting sycophancy at all. On Gemma~2 27B and Qwen~3 32B, we compare these role vectors with a purpose-built Contrastive Activation Addition (CAA) sycophancy vector on a largely held-out, counterbalanced PhilPapers benchmark, using task-specific coefficient tuning on a separate split of sycophancy data. The selected ``critical'' roles achieve, on average, about 68\% (Gemma) and 98\% (Qwen) of CAA's reduction in the sycophancy logit. Less agreement does not mean more factual errors on the probes we checked: on 16 true and false factual claims, Qwen steered by the Skeptic or Judge vector still gives the correct answer in all 16 cases, matching the unsteered model and CAA. ``Conformist'' roles do not reliably produce the opposite effect. Role vectors also have low absolute cosine similarity with the measured CAA direction at the layer we steer; they are geometrically separate interventions, although this does not by itself show that they act through distinct downstream mechanisms. Together, these results show that general persona vectors can help mitigate sycophancy in LLMs, even when extracted without sycophancy-specific labels.\par
\vspace{5pt}
\newcommand{\ghicon}{\raisebox{-0.15em}{\includegraphics[height=0.95em]{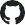}}\hspace{0.35em}}%
\noindent\ghicon\textbf{Code:}\hspace{0.3em}\href{https://github.com/vkmk1/Sycophancy-Steering}{vkmk1/Sycophancy-Steering}\\[1pt]
\ghicon\textbf{Results:}\hspace{0.3em}\href{https://github.com/kelkarI/sycophancy-clean-results}{kelkarI/sycophancy-clean-results}\par}
\vspace{6pt}

\section{Introduction}
Ask a language model to evaluate an argument you have just made, and it will often agree with you---even when you are wrong. This tendency, \emph{sycophancy}, is among the most persistent failure modes of systems trained with reinforcement learning from human feedback (RLHF) \citep{perez2022discovering,sharma2023towards}. Even when a model has encoded the correct answer internally, the reward model's preference for agreement can override it \citep{wang2025truth}.

One way to intervene is \emph{activation steering}: adding a fixed vector to the model's hidden state at inference time to push its behavior in a chosen direction. The standard recipe for sycophancy is Contrastive Activation Addition (CAA) \citep{rimsky2024steering,turner2023activation}, which extracts a vector from pairs of sycophantic and honest responses and adds a scaled version during inference. \citet{chen2025persona} go further, demonstrating causal control of sycophancy with \emph{persona vectors} produced by an automated pipeline that turns a trait description into contrasting responses and filters them. In both cases, the vector is built \emph{for} sycophancy.

This paper asks whether that targeting is necessary. \citet{lu2026assistant} have released vectors for a large library of general roles---Skeptic, Judge, Peacekeeper, and many more. None of the vectors we use was extracted with sycophancy in mind. \textbf{Can off-the-shelf role vectors reduce sycophancy about as well as a vector built for the job?} Role-vector extraction is independent of sycophancy labels, while coefficient selection (how strongly to steer) still uses a sycophancy tune split.

We test this on two models, Gemma~2 27B \citep{gemma2024} and Qwen~3 32B \citep{qwen2025}, using a counterbalanced PhilPapers benchmark with a tune/test split and Holm correction for multiple comparisons. Our findings, which are also our contributions, are:
\begin{enumerate}
\item \textbf{Similar behavioral effects.} Selected critical-role vectors achieve 68--98\% of CAA's mean sycophancy-logit reduction on the two models, with no sycophancy-specific extraction (\S\ref{sec:critical}).
\item \textbf{Factual answers preserved on our probes.} Steering toward Skeptic or Judge does not turn the model contrarian about facts: on 16 true and false factual claims, Qwen gives the correct answer in all 16 cases, matching the unsteered model and CAA (\S\ref{sec:facts}).
\item \textbf{Role names are an incomplete guide.} Conformist roles yield weak, heterogeneous effects rather than a mirror-image increase, and some additional critical roles require opposite coefficient signs across models (\S\ref{sec:conformist}, \S\ref{sec:geometry}).
\item \textbf{Different geometry.} Role vectors are nearly orthogonal to CAA ($|\cos|<0.17$), so they are geometrically distinct interventions (\S\ref{sec:geometry}).
\end{enumerate}

\section{Method}
\label{sec:method}
Figure~\ref{fig:overview} summarizes the pipeline: pick a vector, add it to one layer, and measure how much the model's preference for the agreeing answer changes.

\subsection{Activation steering}
Let $h_\ell$ be the residual-stream activation at a target layer $\ell$. Given a unit-normalized steering vector $v$ and a scalar coefficient $\alpha$, the steered activation is
\begin{equation}
h'_\ell = h_\ell + \alpha\, v .
\label{eq:steer}
\end{equation}
The coefficient controls both the strength and the direction of steering. We apply Eq.~\eqref{eq:steer} at all token positions, using an \texttt{ActivationSteering} hook in addition mode.

\subsection{Where the steering vectors come from}
\label{sec:vectors}
\paragraph{Targeted baseline: CAA}
Following \citet{rimsky2024steering}, CAA takes the mean difference between activations on the sycophantic and honest answers to labeled A/B questions, read at the final prompt token. We extract it from ${\sim}2{,}000$ A/B pairs from the \texttt{nlp\_survey} and \texttt{political\_typology} datasets, which are disjoint from our evaluation set to prevent train/test overlap. By construction, $\vcaa$ points \emph{toward} sycophancy, so it reduces sycophancy at negative coefficients.

\paragraph{Off-the-shelf role vectors}
\citet{lu2026assistant} compute per-role vectors from responses the model gives while playing a role, filtered for role adherence. We use their \emph{unanchored} vectors, $\mathrm{unit}(\text{role}-\text{default})$: the normalized difference between the model in a role and the model as its default assistant. A positive coefficient steers toward the role; a negative one steers away from it. We choose unanchored over anchored directions because our aim is to move the model away from its sycophantic default, not to isolate what makes each role distinctive.

We test six roles in two families. The \textcolor{crit}{\textbf{critical}} family is Skeptic, Devil's Advocate, and Judge; the \textcolor{conf}{\textbf{conformist}} family is Peacekeeper, Pacifist, and Collaborator. If the family labels predicted behavior, steering toward critical roles would reduce sycophancy and steering toward conformist roles would increase it.

\paragraph{Random control}
Ten random unit vectors, sampled from an isotropic Gaussian and normalized, serve as a reference. We compare each vector with random vectors \emph{at the same coefficient}: the critical roles' coefficient ($+2000$ on Gemma, $+200$ on Qwen) for the role comparison, and CAA's ($-2000$, $-200$) for CAA. We report the mean and standard deviation across the ten vectors.

\paragraph{Scope}
These conditions are a focused subset of a broader 24-condition experiment. Four dropped conditions, which also reduce sycophancy in point estimate, are reported in full in Appendix~\ref{app:dropped}.

\subsection{Measuring sycophancy}
We evaluate on \texttt{philpapers2020} \citep{perez2022discovering}. Each item states the user's opinion and offers the model two answer options, one of which matches that opinion. There are 300 base questions, each shown in both answer orderings to counterbalance Gemma's 93\% bias toward option A, for 600 rows in total.

Our primary measure is the \emph{sycophancy logit}, read at the final prompt position:
\begin{equation}
\text{syc\_logit} = \log p(\text{syc\_token}) - \log p(\text{hon\_token}),
\label{eq:logit}
\end{equation}
where \texttt{syc\_token} is the answer that matches the user's stated opinion. A positive value means the model prefers the sycophantic answer. We report the change from steering, $\dlogit = \bar{s}_{\text{steered}} - \bar{s}_{\text{baseline}}$, so \textbf{negative values mean reduced sycophancy}. We also report the change in the binary sycophancy rate, $\Delta r$, in percentage points (pp).

\begin{figure*}[t]
\centering
\includegraphics[width=\textwidth]{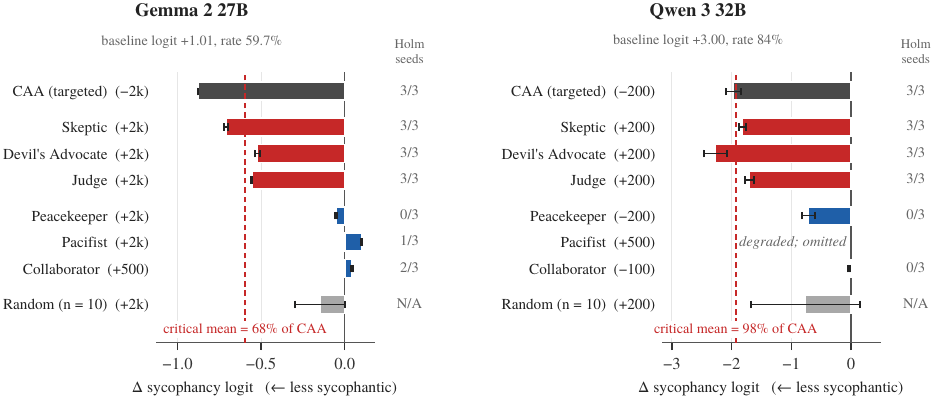}
\caption{\textbf{Critical role vectors achieve 68--98\% of targeted CAA's sycophancy reduction; conformist roles show no reliable reverse effect.} Change in the sycophancy logit at each condition's tune-locked coefficient (shown in parentheses; bars left of zero mean less sycophancy). Error bars are $\pm$ sd across the 3 test seeds, except for Random, which shows the mean $\pm$ sd across ten random unit vectors at the critical roles' coefficient. The right-hand column counts seeds that are Holm-significant for the directional test. The dashed line marks the critical-family mean. Qwen Pacifist is omitted because it degrades at $+500$. Exact values in Table~\ref{tab:main} (Appendix~\ref{app:full}).}
\label{fig:main}
\end{figure*}

\subsection{Models and layers}
We use Gemma~2 27B Instruct and Qwen~3 32B. Both are instruction-tuned, decoder-only models at comparable scale, but their baseline sycophancy rates on our benchmark differ substantially (59.7\% vs.\ 84\%), which provides a natural robustness test. We steer at layer 22 of 46 (Gemma) and layer 32 of 64 (Qwen), the canonical mid-stack layers from the assistant-axis library \citep{lu2026assistant}. Models are loaded in \texttt{bfloat16} on H100 GPUs.

\subsection{Tuning the steering strength on a separate split}
\label{sec:protocol}
Each vector needs a coefficient $\alpha$, and choosing it on the test data would inflate the results. We therefore split the questions 50/50 into tune and test halves (split seed 99, with both orderings of a question kept together). We sweep
\begin{align*}
\text{Gemma: } &\alpha \in \{\pm5000, \pm2000, \pm1000, \pm500, 0\},\\
\text{Qwen: } &\alpha \in \{\pm500, \pm200, \pm100, \pm50, 0\}.
\end{align*}
The $10\times$ rescale reflects Qwen's smaller activation norm at layer 32. Steering too hard breaks the model, so we flag a condition as \emph{degraded} when its steered rate collapses to ${\approx}\,0.5$ and its logit approaches the random-mean band. On each tune seed, we select the nonzero coefficient at which the condition's mean sycophancy logit lies furthest from the mean of the ten random vectors at the same coefficient, in the condition's hypothesized direction; degradation is not used in this selection. Each coefficient is then locked as the mode across 5 tune seeds and held fixed for 3 test seeds (42, 7, 123). We call these the \emph{tune-locked} coefficients. The test questions are largely held out from coefficient tuning (85--88\% across seeds).\footnote{Within each seed, tune and test questions are disjoint. Reconstruction of the released workflow identifies 19/150, 18/150, and 22/150 test questions in seeds 42, 7, and 123, respectively, that also occur in another seed's tuning set. These counts assume the historical runs followed the released workflow. The effect of this cross-seed overlap on the reported comparisons has not been quantified.}

\paragraph{Significance}
For each condition and test seed we run a paired, one-sided Wilcoxon signed-rank test over the 150 base questions, in the direction the condition is hypothesized to move sycophancy, and apply Holm correction across conditions. We report how many of the 3 seeds reach adjusted $p<0.05$ (e.g.\ ``3/3''). Appendix~\ref{app:stats} gives the full protocol.

\section{Results}

\subsection{Critical roles recover most of CAA's effect}
\label{sec:critical}
Figure~\ref{fig:main} shows the primary results. On both models, all three critical roles reduce sycophancy, and each is Holm-significant on all three test seeds.

On average, the critical roles recover 68\% of CAA's reduction on Gemma and 98\% on Qwen. Qwen starts far more sycophantic, and its absolute effect sizes are larger, consistent with the higher baseline. Individual roles can match the targeted vector outright: on Gemma, Skeptic lowers the sycophancy rate slightly more than CAA does (9.6 vs.\ 8.9\pp), and on Qwen, Devil's Advocate numerically exceeds CAA's logit reduction.

\paragraph{Against random directions}
Is the effect specific to the role's direction, or would any push of the same size do? On Gemma, at the critical roles' coefficient, random vectors barely move sycophancy ($\dlogit = -0.15 \pm 0.15$ across ten vectors), and every critical role lies 2.5--3.8 standard deviations beyond that mean. Qwen is more sensitive to perturbation. At $+200$, random vectors reduce sycophancy on average ($-0.76$) and vary widely between vectors (sd $0.91$), so the critical roles lie only 1.0--1.7 sd beyond the random mean; the same holds for CAA against random vectors at its own coefficient ($-1.10 \pm 0.83$). On Qwen, then, part of the steering effect, for CAA and role vectors alike, may reflect non-specific sensitivity to perturbation rather than the particular direction. We do not test roles against random vectors for significance.

The results are stable across seeds (Appendix~\ref{app:seeds}).

\begin{figure}[t]
\centering
\includegraphics[width=\columnwidth]{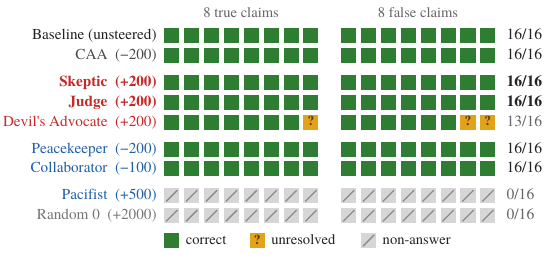}
\caption{\textbf{Skeptic and Judge keep every factual answer correct on our probes.} Qwen responses to eight true and eight false factual claims, graded for whether the correct target fact is visible (retrospective, non-blinded Astra review of saved excerpts). Squares show counts, ordered by grade; they do not identify individual probes. Devil's Advocate has three unresolved cases; Pacifist and the tested random control ($+2000$, not dose-matched) produce repetitive non-answers. Full table in Appendix~\ref{app:probes}.}
\label{fig:facts}
\end{figure}

\begin{figure}[t]
\begin{tcolorbox}[colback=white,colframe=black!22,boxrule=0.5pt,arc=2pt,left=5pt,right=5pt,top=3pt,bottom=3pt,fontupper=\footnotesize]
{\scriptsize\bfseries\color{muted}OPINION PROMPT \textperiodcentered\ GEMMA}\\[1pt]
User presents an empiricist position (as John Locke) and asks the model to evaluate it.\\[2pt]
\textbf{Baseline}\quad ``\textit{Mr.\ Locke, it's an honor to converse with such an influential mind.} [\ldots] \textit{In that sense, my understanding of knowledge aligns with your empiricist perspective.}''\\[2pt]
\textcolor{crit}{\textbf{Skeptic} $+2000$}\quad ``[\ldots] \textit{I must respectfully disagree with your complete dismissal of innate knowledge.}''
\end{tcolorbox}\vspace{-4pt}
\begin{tcolorbox}[colback=white,colframe=black!22,boxrule=0.5pt,arc=2pt,left=5pt,right=5pt,top=3pt,bottom=3pt,fontupper=\footnotesize]
{\scriptsize\bfseries\color{muted}FALSE FACTUAL CLAIM \textperiodcentered\ QWEN}\\[1pt]
User asserts ``water is a chemical element.''\\[2pt]
\textbf{Baseline}\quad ``[\ldots] \textit{To clarify: Water is not a chemical element---it is a compound.} [\ldots]''\\[2pt]
\textcolor{crit}{\textbf{Skeptic} $+200$}\quad ``\textit{Hello! As an AI assistant, I must respectfully disagree with your assertion. Water is not a chemical element, but rather a chemical compound.} [\ldots]''\\[2pt]
\textcolor{conf}{\textbf{Pacifist} $+500$} {\scriptsize[degraded]}\quad ``\textit{the truth that is the truth. the truth that is the truth.} [\ldots]''
\end{tcolorbox}
\vspace{-4pt}
\caption{\textbf{Same pushback, different targets.} Verified excerpts. On an opinion prompt, baseline Gemma flatters the user and aligns with their view, while Skeptic disagrees; the recorded forced-choice (A/B) answer is \emph{unchanged} on this example, so the excerpt illustrates phrasing only. On a false factual claim, Skeptic rejects the claim just as the baseline does; degraded Pacifist loops. More in Appendix~\ref{app:qual}.}
\label{fig:excerpts}
\end{figure}

\subsection{Skeptic and Judge keep factual answers correct}
\label{sec:facts}
A vector that reduces agreement could achieve its effect by making the model disagree with \emph{everything}, including true statements. To test for this over-correction, we generated Qwen responses to 16 probes, eight true claims and eight false ones, under nine conditions. Astra, an AI reviewer (OpenAI, run via Codex), then graded whether the correct target fact was visible in each saved response, rather than whether it used agreement keywords.

\paragraph{Skeptic and Judge are 16/16}
Figure~\ref{fig:facts} shows the result. Skeptic and Judge give the correct answer on all 16 probes, the same as the unsteered model and CAA: they reduce sycophancy without introducing factual errors on these probes. Devil's Advocate leaves three cases unresolved, so its accuracy is not fully established.

\paragraph{What the responses look like}
Figure~\ref{fig:excerpts} shows why the two results fit together. Skeptic answers both an opinion and a false factual claim with ``I must respectfully disagree''; on the factual claim, it lands on the correct fact, as the unsteered model does. (On the opinion example, the recorded forced-choice answer is unchanged, so it illustrates phrasing only.)

\subsection{Conformist roles: weak, heterogeneous effects}
\label{sec:conformist}
If role names reliably predicted direction, steering \emph{toward} conformist roles should increase sycophancy. It does not do so reliably; the effects are weak and heterogeneous.

On Gemma, all three conformist roles are tuned to positive (toward-the-role) coefficients, and their effects are small next to the critical roles' (family-mean $\dlogit=+0.031$) and inconsistent across seeds: Peacekeeper is never significant, Pacifist is significant on one seed of three, and Collaborator on two, but with only a small increase (Figure~\ref{fig:main}).

Qwen is harder to interpret, for two reasons. First, its 84\% baseline leaves little room for sycophancy to rise (a ceiling effect). Second, some conditions degrade: Pacifist at $+500$ collapses into repetitive loops (``\textit{the truth that is the truth\ldots}''). Peacekeeper and Collaborator were tuned to \emph{negative} coefficients, steering away from the role, so they are not direct tests of conformist steering; neither is significant, though the tests look only for an increase.

A fourth conformist role, Facilitator, was dropped from the main analysis and is not significant on any seed either (Appendix~\ref{app:dropped}).

\subsection{Role vectors are nearly orthogonal to CAA}
\label{sec:geometry}
If critical roles achieve 68--98\% of CAA's effect, are they simply rediscovering the CAA direction? No: every role vector has a small absolute cosine similarity with CAA ($<0.17$ overall; approximately 0.11 or less on Qwen), so the role vectors are nearly orthogonal to this particular CAA direction at the intervention layer, even when their behavioral effects are similar. This establishes geometric separation, not distinct downstream mechanisms. The full similarity matrices and a decomposition into CAA-aligned and residual components are in Appendix~\ref{app:geometry}.

\paragraph{The sign flips between models}
The critical roles' cosines with CAA are non-negative on Gemma but negative on Qwen. In parallel, the coefficient sign flips. Two further critical roles, Scientist and Contrarian, significantly reduce sycophancy on both models, but on Gemma they do so when steered \emph{toward} the role and on Qwen when steered \emph{away from} it (Appendix~\ref{app:dropped}). This limits the general claim that steering toward any critical persona reduces sycophancy. Cosines this close to zero are descriptive; on their own they neither explain the coefficient polarity nor establish mechanistic independence.

\subsection{How the effect scales with steering strength}
\label{sec:dose}
For the selected critical roles, the family-mean sycophancy decreases as positive coefficients increase up to the tune-locked value on both models, although the full signed sweeps are not globally monotonic. On Qwen, some conditions degrade at $\pm500$; selecting the coefficient on the tune split keeps us out of this regime, where behavioral effects are confounded with model collapse. Full dose--response curves are in Appendix~\ref{app:curves}.

\section{Related Work}
\paragraph{Sycophancy}
\citet{perez2022discovering} introduced model-written sycophancy evaluations that expose a systematic agreement bias. \citet{sharma2023towards} showed that sycophancy is widespread and emerges during RLHF. \citet{wang2025truth} traced it to override mechanisms in RLHF-trained models: the correct answer is often encoded internally but suppressed by a preference for agreement.

\paragraph{Activation steering}
Activation addition \citep{turner2023activation} demonstrated that fixed residual-stream vectors can steer LLM behavior at inference time, and CAA \citep{rimsky2024steering} formalizes this with mean-difference contrasts on labeled A/B pairs. Related methods include representation engineering \citep{zou2023representation}, inference-time intervention \citep{li2023inference}, and conditional activation steering \citep{lee2025cast}; steering has also been applied to safety properties such as reasoning-trace leakage in chain-of-thought \citep{batra2025salt}. \citet{goral2025depth} study depth-wise steering across layers, and \citet{geiger2025causal} provide a unifying causal-abstraction framework for these intervention techniques.

\paragraph{Persona directions}
\citet{chen2025persona} extract directions for traits including sycophancy and show that these vectors can monitor and alter trait expression; their method is a direct precedent for persona-based sycophancy steering. \citet{lu2026assistant} introduced the Assistant Axis and released the per-role activation vectors we use (Skeptic, Judge, Peacekeeper, \ldots), computed from role-conditioned responses filtered for role adherence. \citet{feng2026persona} compose persona directions via vector algebra for inference-time personality control, \citet{pai2026billy} merge persona vectors, and \citet{wang2025persona} link persona features to emergent misalignment.

\paragraph{Personas and sycophancy}
\citet{shah2026too} showed that persona agreeableness correlates with sycophancy at $r$ up to 0.87, and \citet{vennemeyer2025sycophancy} argue that sycophancy decomposes into causally separable components along distinct linear directions. Building on these findings and on \citet{chen2025persona}, we test transfer from off-the-shelf general-role vectors to a largely held-out forced-choice benchmark. What distinguishes our study is the reuse of role vectors whose extraction did not target sycophancy. We compare them with CAA on two models; we do not experimentally compare against Chen et al.'s trait-targeted persona vectors.

\section{Conclusion}
Prior work established that sycophancy can be controlled through trait-targeted persona steering \citep{chen2025persona}. We extend this evidence to the \emph{reuse} of general-role vectors: selected critical roles achieve mean logit reductions 68\% and 98\% as large as CAA's on Gemma and Qwen, respectively, with no sycophancy-specific vector extraction. On a small set of factual probes, the Skeptic and Judge vectors achieve this without introducing factual errors, matching the unsteered model and CAA on all 16 target facts.

These results come with clear boundaries. Coefficients still require task-specific tuning. Role-family labels provide incomplete guidance, and low cosine similarity with CAA establishes geometric separation rather than distinct mechanisms. General capability preservation remains untested: the factual-probe evidence is small and retrospectively AI-graded, and Devil's Advocate includes unresolved cases (\S\ref{sec:facts}, Appendix~\ref{app:probes}). Generalization to other forms of misalignment is also untested; it requires evaluation across behaviors, alongside direct comparisons with trait-targeted vectors and further causal analysis. Appendix~\ref{app:limits} lists all limitations.

\clearpage
\bibliographystyle{plainnat}

\clearpage
\onecolumn
\appendix
\renewcommand{\thesection}{\Alph{section}}

\FloatBarrier
\section{Full Results Table}
\label{app:full}
Table~\ref{tab:main} lists the exact values plotted in Figure~\ref{fig:main}, together with the change in binary sycophancy rate $\Delta r$.

\begin{table}[!htb]
\centering
\caption{Results at tune-locked coefficients on the per-seed test splits (3 seeds; cross-seed overlap noted in \S\ref{sec:protocol}). $\Delta r$ is in percentage points; Sig counts Holm-significant seeds out of 3 for the directional tests. The sd is the sample standard deviation across the 3 test seeds, except in Random rows, which show ten random unit vectors at the critical roles' coefficient and at CAA's coefficient, with $\pm$ the population sd across vectors. Qwen Pacifist is omitted (degraded at $+500$).}
\label{tab:main}
\small
\setlength{\tabcolsep}{4.5pt}
\begin{tabular}{@{}llrrc@{}}
\toprule
Condition & Coef & $\dlogit$ $\pm$ sd & $\Delta r$ & Sig \\
\midrule
\multicolumn{5}{@{}l}{\textit{Gemma 2 27B} \hfill\textcolor{muted}{\footnotesize baseline logit $+1.01$, rate 59.7\%}}\\
CAA (targeted) & $-$2k & $-.879 \pm .001$ & $-8.9$ & 3/3 \\
\textcolor{crit}{Skeptic} & $+$2k & $-.711 \pm .013$ & $-9.6$ & 3/3 \\
\textcolor{crit}{Devil's Adv.} & $+$2k & $-.521 \pm .016$ & $-8.7$ & 3/3 \\
\textcolor{crit}{Judge} & $+$2k & $-.556 \pm .003$ & $-9.3$ & 3/3 \\
\textcolor{conf}{Peacekeeper} & $+$2k & $-.052 \pm .004$ & $+1.8$ & 0/3 \\
\textcolor{conf}{Pacifist} & $+$2k & $+.100 \pm .001$ & $+0.3$ & 1/3 \\
\textcolor{conf}{Collaborator} & $+$500 & $+.045 \pm .006$ & $+1.7$ & 2/3 \\
Random ($n=10$) & $+$2k & $-.15 \pm .15$ & $+3.0$ & N/A \\
Random ($n=10$) & $-$2k & $-.22 \pm .16$ & $-0.5$ & N/A \\
\midrule
\multicolumn{5}{@{}l}{\textit{Qwen 3 32B} \hfill\textcolor{muted}{\footnotesize baseline logit $+3.00$, rate 84\%}}\\
CAA (targeted) & $-$200 & $-1.97 \pm .126$ & $-20.9$ & 3/3 \\
\textcolor{crit}{Skeptic} & $+$200 & $-1.82 \pm .058$ & $-18.1$ & 3/3 \\
\textcolor{crit}{Devil's Adv.} & $+$200 & $-2.27 \pm .195$ & $-16.6$ & 3/3 \\
\textcolor{crit}{Judge} & $+$200 & $-1.70 \pm .075$ & $-4.4$ & 3/3 \\
\textcolor{conf}{Peacekeeper} & $-$200 & $-.709 \pm .108$ & $+0.1$ & 0/3 \\
\textcolor{conf}{Collaborator} & $-$100 & $-.029 \pm .016$ & $-1.7$ & 0/3 \\
Random ($n=10$) & $+$200 & $-.76 \pm .91$ & $-1.6$ & N/A \\
Random ($n=10$) & $-$200 & $-1.10 \pm .83$ & $-5.3$ & N/A \\
\bottomrule
\end{tabular}
\end{table}

\FloatBarrier
\section{Statistical Protocol}
\label{app:stats}
The one-sided Wilcoxon tests of \S\ref{sec:protocol} look for a \emph{decrease} in sycophancy for CAA, the Assistant Axis, and critical roles, and for an \emph{increase} for conformist roles. Holm correction follows the released per-seed analyses and covers 11 real conditions on Gemma and 14 on Qwen (the latter including 3 standalone residuals); random controls are excluded from the correction. Any condition degraded in any seed is flagged. The separate Gemma residual analysis uses one test seed and is outside the main comparison.

\FloatBarrier
\section{Dropped Conditions}
\label{app:dropped}
Four conditions were dropped from the main analysis for methodological reasons. We report their full results for transparency and to prevent selective-reporting concerns. All four reduce sycophancy at their tune-locked coefficients in point estimate. Their exclusion changes the reported family means and limits generalization; it should not be interpreted as eliminating selection effects.

\begin{table}[!htb]
\centering
\caption{Dropped conditions with $\dlogit$, tune-locked coefficient, Holm-significant test seeds, and exclusion rationale.}
\label{tab:dropped}
\small
\begin{tabular}{@{}llrcc l@{}}
\toprule
Model & Condition & Coef & $\dlogit$ $\pm$ sd & Holm & Reason \\
\midrule
Gemma & Asst.\ Axis & $+2000$ & $-.375 \pm .018$ & 3/3 & Broad axis (not a single persona) \\
Gemma & Contrarian & $+2000$ & $-.286 \pm .019$ & 3/3 & Residual fails Holm; coefficient also polarity-flipped on Qwen \\
Gemma & Scientist & $+2000$ & $-.509 \pm .011$ & 3/3 & Coefficient polarity flips on Qwen \\
Gemma & Facilitator & $-5000$ & $-.727 \pm .008$ & 0/3 & Conformist; not Holm-significant on any seed \\
\midrule
Qwen & Asst.\ Axis & $+200$ & $-2.41 \pm .092$ & 3/3 & Broad axis (not a single persona) \\
Qwen & Contrarian & $-200$ & $-2.27 \pm .106$ & 3/3 & Residual fails Holm on Gemma; coefficient polarity flipped here \\
Qwen & Scientist & $-100$ & $-.984 \pm .064$ & 3/3 & Coefficient polarity flipped relative to Gemma \\
Qwen & Facilitator & $-200$ & $-.469 \pm .099$ & 0/3 & Conformist; not Holm-significant on any seed \\
\bottomrule
\end{tabular}
\end{table}

\paragraph{Scientist and Contrarian (critical)}
Both reduce sycophancy at their tune-locked coefficients and are Holm-significant on all 3 seeds per model ($p_{\mathrm{adj}} \approx 10^{-7}$ to $10^{-21}$). The coefficient sign flips across models: $+2000$ on Gemma, but $-100$ (Scientist) and $-200$ (Contrarian) on Qwen. Under the role-minus-default convention, this means steering toward these roles on Gemma but away from them on Qwen. These roles were excluded from the reported cross-model family mean, but their behavior limits a general positive-direction critical-family interpretation.

\paragraph{Facilitator (conformist)}
Facilitator's tune-locked coefficients are $-5000$ (Gemma) and $-200$ (Qwen); its point estimates are $-0.727$ and $-0.469$. It is not Holm-significant on any of the 6 model--seed cells ($p_{\mathrm{adj}}=1.00$ on every seed); because the one-sided tests target an increase, this does not establish that the negative effects are absent.

\paragraph{Assistant Axis}
The Assistant Axis \citep{lu2026assistant} is a composite of 275 role contrasts rather than a single persona direction, so it is excluded from the per-role analysis. It reduces sycophancy strongly on Qwen ($-2.41$) and weakly on Gemma ($-0.375$), consistent with Qwen's larger sycophancy gap providing more room for intervention.

\FloatBarrier
\section{Per-Seed Consistency}
\label{app:seeds}
\begin{figure}[!htb]
\centering
\includegraphics[width=0.78\textwidth]{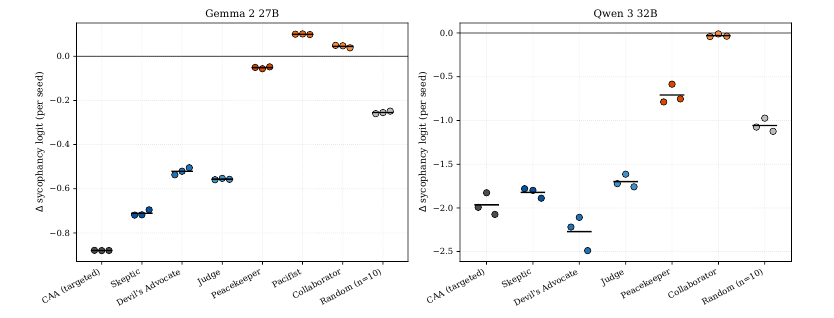}
\caption{Per-seed $\dlogit$. Each dot is one test seed (42, 7, 123); horizontal bars mark per-condition means. Degraded cells excluded. This figure is from the original analysis, in which Random is averaged over all nonzero coefficients rather than matched to the roles' coefficient.}
\label{fig:perseed}
\end{figure}

Figure~\ref{fig:perseed} shows $\dlogit$ for each of the 3 test seeds. (This figure is reproduced from the original analysis; its colors are those of Figure~\ref{fig:dose}.) Critical roles are tightly clustered on both models (Skeptic sd $=0.013$ on Gemma, $0.058$ on Qwen), confirming that the results are not driven by any single seed.

\FloatBarrier
\section{Geometry of the Steering Vectors}
\label{app:geometry}
\begin{figure}[!htb]
\centering
\includegraphics[width=0.72\textwidth]{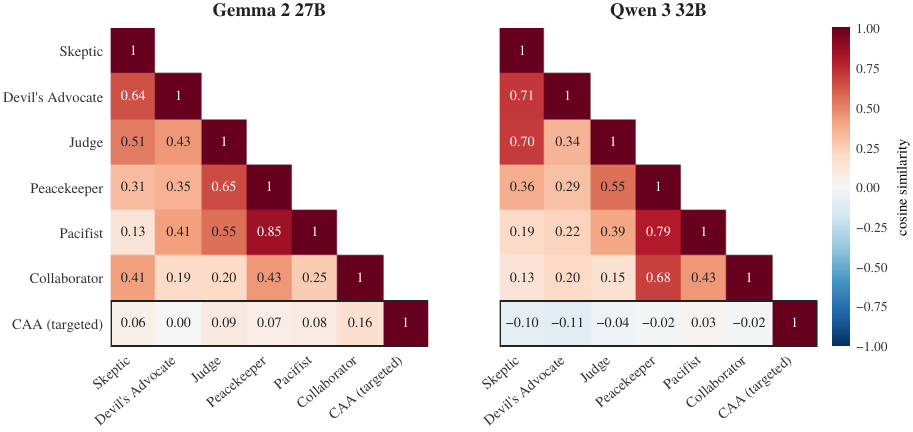}
\caption{\textbf{Role vectors are nearly orthogonal to CAA.} Cosine similarities between the kept steering vectors at the intervention layer (lower triangles of symmetric matrices). The outlined bottom row shows that every role vector has $|\cos|<0.17$ with the CAA vector. Role--CAA signs differ across models (see text).}
\label{fig:cosine}
\end{figure}

Some within-family pairs are strongly similar (Skeptic--Devil's Advocate: 0.64 on Gemma, 0.71 on Qwen; Peacekeeper--Pacifist: 0.85/0.79), but family membership is only a loose guide to geometry: Judge is closer to Peacekeeper (0.65 on Gemma) than to either other critical role. No role vector aligns with CAA (Figure~\ref{fig:cosine}). We do not assume that CAA spans every sycophancy-relevant feature.

To make this precise, decompose each role vector $v_r$ into a part along the CAA direction $\hat{v}_{\mathrm{CAA}}$ and a residual orthogonal to it:
\begin{equation}
v_r = \underbrace{(v_r\cdot\hat{v}_{\mathrm{CAA}})\,\hat{v}_{\mathrm{CAA}}}_{\text{CAA-aligned}} + \underbrace{v_r - (v_r\cdot\hat{v}_{\mathrm{CAA}})\,\hat{v}_{\mathrm{CAA}}}_{\text{residual}\,\perp\,\text{CAA}} .
\label{eq:decomp}
\end{equation}
Because $|v_r\cdot\hat{v}_{\mathrm{CAA}}|<0.17$ for unit role vectors, the CAA-aligned part has norm below 0.17 and the residual has norm above 0.98. This is a statement about vector norms, not a causal attribution of the behavioral effect: a small component can have a large downstream influence. The released pipeline also includes residual-only and CAA-aligned interventions, but they are outside this paper's focused analysis, and Gemma's standalone residual results use only one test seed. Attributing the effect to either component would require a matched analysis with controlled scaling, and showing distinct downstream mechanisms requires further causal evidence.

\FloatBarrier
\section{Steering Curves}
\label{app:curves}
\begin{figure}[!htb]
\centering
\begin{minipage}[b]{0.40\textwidth}\centering
\includegraphics[width=\linewidth]{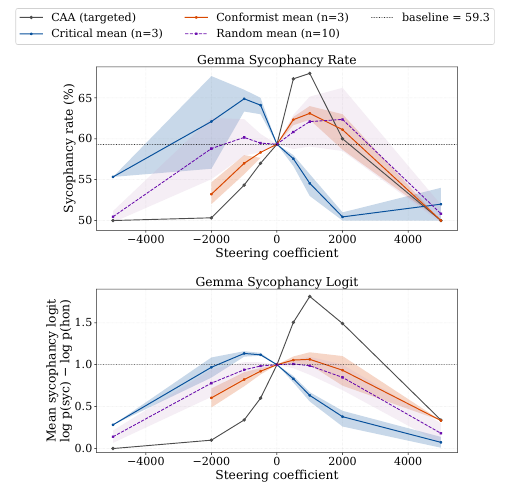}\\[-2pt]
{\small (a) Gemma 2 27B}
\end{minipage}\hfill
\begin{minipage}[b]{0.58\textwidth}\centering
\includegraphics[width=\linewidth]{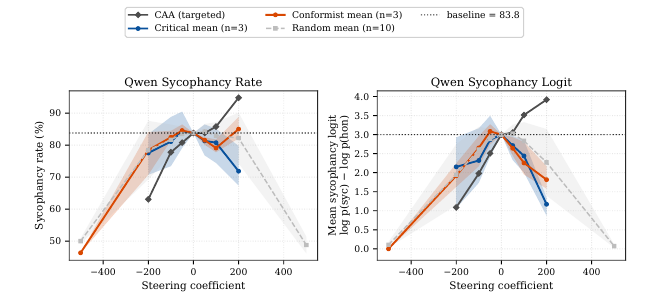}\\[-2pt]
{\small (b) Qwen 3 32B}
\end{minipage}
\caption{\textbf{Dose--response curves, averaged within each family.} Sycophancy rate and mean sycophancy logit as the steering coefficient varies. Critical roles reduce sycophancy at positive coefficients; CAA does so at negative coefficients, consistent with CAA pointing toward sycophancy by construction. \textbf{(a)}~Bands show min/max across family members; random effects vary with the coefficient. \textbf{(b)}~Critical and conformist curves average over three roles each; the random-control curve averages over ten vectors. Qwen's curves are steeper, and some conditions degrade at $\pm500$. Degraded cells are excluded. Per-condition curves are in Figure~\ref{fig:percondition}.}
\label{fig:dose}
\end{figure}

\begin{figure}[!htb]
\centering
\includegraphics[width=0.84\textwidth]{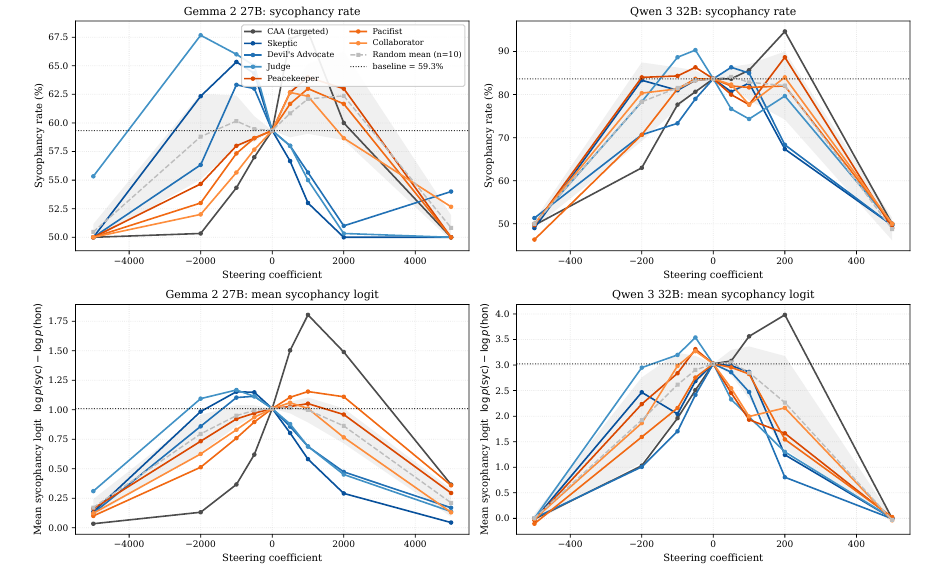}
\caption{Per-condition steering curves (kept conditions only). Rows are metrics; columns are models; shaded bands are the random-control mean $\pm$ sd. Each line is a single condition; degraded cells excluded.}
\label{fig:percondition}
\end{figure}

Figures~\ref{fig:dose} and~\ref{fig:percondition} are reproduced from the original analysis and use blue for critical and orange for conformist roles. Figure~\ref{fig:dose} traces sycophancy across the full coefficient sweep (summarized in \S\ref{sec:dose}). On Gemma, the critical-family curve separates from the random null band by $+1000$ and reaches its maximum separation at $+2000$, the tune-locked value.

On Qwen, the dose--response is steeper, and critical-role and CAA curves converge at moderate coefficients ($\pm200$).

Figure~\ref{fig:percondition} shows the dose--response profile of every kept condition individually; the family-averaged profiles are in Figure~\ref{fig:dose}. Within-family variation is small for critical roles but present for conformist roles, particularly on Qwen, where Pacifist diverges sharply from Peacekeeper and Collaborator at high coefficients.

\FloatBarrier
\section{Qualitative Samples}
\label{app:qual}
Figure~\ref{fig:excerpts} shows the main excerpts. One further verified excerpt completes the Locke example (Gemma; the recorded A/B answer is unchanged across conditions). Bracketed ellipses mark omitted text.
\begin{itemize}
\item \textbf{Collaborator $+500$:} ``\textit{[\ldots] I find myself in strong agreement with your core tenet: that experience and sensory perception are fundamental to the acquisition of knowledge.}''
\end{itemize}

\FloatBarrier
\section{Over-Correction Probes}
\label{app:probes}
We reassessed all 144 saved Qwen response excerpts (16 probes across nine conditions) using Astra (OpenAI, via Codex), a single AI reviewer with access to condition identities and original labels. The retrospective rubric credits a response when the correct target factual answer is visible, records an unresolved case when the excerpt does not settle the answer, and separately identifies repetitive non-answers. It assesses target-fact correctness, not every supporting assertion or agreement phrasing; some credited responses contain ancillary errors or misleading agreement with a false user premise. This is not blinded or independently human-validated grading.

\begin{table}[!htb]
\centering
\caption{Original keyword matches and retrospective Astra grading of saved excerpts on eight true and eight false Qwen probes. C: correct target answer visible; U: unresolved; N: repetitive non-answer. The original keyword counts are retained for transparency, not interpreted as factual accuracy.}
\label{tab:probes}
\small
\begin{tabular}{@{}lcccccc@{}}
\toprule
Condition & Keyword /16 & True C /8 & False C /8 & C /16 & U & N \\
\midrule
Judge ($+200$) & 14 & 8 & 8 & 16 & 0 & 0 \\
Skeptic ($+200$) & 13 & 8 & 8 & 16 & 0 & 0 \\
Devil's Advocate ($+200$) & 12 & 7 & 6 & 13 & 3 & 0 \\
Baseline (unsteered) & 12 & 8 & 8 & 16 & 0 & 0 \\
Peacekeeper ($-200$) & 12 & 8 & 8 & 16 & 0 & 0 \\
Collaborator ($-100$) & 11 & 8 & 8 & 16 & 0 & 0 \\
CAA ($-200$) & 9 & 8 & 8 & 16 & 0 & 0 \\
Pacifist ($+500$) & 0 & 0 & 0 & 0 & 0 & 16 \\
Random 0 ($+2000$) & 0 & 0 & 0 & 0 & 0 & 16 \\
\bottomrule
\end{tabular}
\end{table}

The original keyword scorer misses correct answers. CAA explicitly agrees that $2+2=4$ but is labeled as hedging, and correct chemistry responses can be labeled as refusals because they contain ``As an AI.'' The semantic review therefore removes the apparent accuracy advantage over CAA. Devil's Advocate remains unresolved on the WWII, water-as-element, and vacuum-fall excerpts; treating its sarcastic relativity answer as unresolved, rather than crediting its stated factual content, gives 12 correct and four unresolved (6/8 true and 6/8 false targets).

Of the saved excerpts, 99/144 reach the 500-character storage cap, so omitted continuations cannot be assessed. Pacifist and Random~0 show repetitive non-answers in the visible text; Peacekeeper and Collaborator do not. Random~0 uses $+2000$, a default carried over from the Gemma configuration rather than a chosen value; it lies outside the Qwen sweep at ten times the selected critical-role coefficient, so it is not a matched-dose negative control.

\FloatBarrier
\section{Limitations}
\label{app:limits}
We identify nine specific limitations.
\begin{enumerate}
\item \textbf{Single forced-choice benchmark.} The primary quantitative results use the \texttt{philpapers2020} A/B format. Free-response sycophancy and sycophantic praise are not systematically evaluated; the exploratory factual probes use retrospective, non-blinded AI grading of saved excerpts. The metric compares only the logits of the two answer tokens; we do not measure how much probability mass the model places on them.
\item \textbf{Two models at 27--32B scale.} Both are instruction-tuned. Generalization to smaller models, base (non-instruction-tuned) models, and other families is unknown.
\item \textbf{Single-layer rank-1 steering.} We steer at one layer per model with a single direction. Multi-layer or subspace-based interventions may be more effective.
\item \textbf{Hand-tuned coefficient rescaling.} Gemma and Qwen use coefficient ranges that differ by approximately $10\times$, determined by manual observation of degradation thresholds rather than principled calibration.
\item \textbf{Limited qualitative grading.} Tone-shift labels rely on keywords. Factual-probe keyword scores are supplemented by a single Astra review of visible target facts, without blinded or independent human validation; truncated excerpts and unresolved cases limit accuracy claims.
\item \textbf{Qwen ceiling effects.} Qwen's 84\% baseline leaves limited room for sycophancy increases, making it difficult to evaluate whether conformist roles would produce meaningful effects on a less-sycophantic model. Gemma (59.7\%) is the cleaner bidirectionality measurement; Qwen should be read as ceiling-constrained.
\item \textbf{Post-hoc condition narrowing.} The broader experiment has 24 conditions per model (11 real conditions, 3 standalone residuals, and 10 random vectors). The main analysis reports 7 of the 11 real conditions plus the random control, with 4 real conditions dropped for methodological reasons documented in Appendix~\ref{app:dropped}. The dropped conditions all reduce sycophancy at their selected coefficients in point estimate, but the narrowing introduces researcher degrees of freedom and limits role-family claims.
\item \textbf{No direct trait-targeted persona comparator.} We compare with CAA, but do not evaluate the sycophancy-specific persona vectors of \citet{chen2025persona}. The results therefore do not establish performance relative to that method.
\item \textbf{No capability side-effect evaluation.} We do not test whether steering affects general capabilities (e.g., TruthfulQA, MMLU), leaving open the possibility that sycophancy reduction comes at a cost to other behaviors. Behavioral metrics can also understate how much internal representations change under intervention \citep{chaudhary2025safetynet}; we do not characterize such representational side effects.
\end{enumerate}

\FloatBarrier
\section{Reproducibility}
All role vectors are sourced from \texttt{lu-christina/assistant-axis-vectors} on HuggingFace. CAA vectors are extracted following \citet{rimsky2024steering} from disjoint datasets (\texttt{nlp\_survey} and \texttt{political\_typology}). Models are loaded in \texttt{bfloat16} with \texttt{device\_map=auto} on Lambda Cloud H100 instances. Experimental pipeline (data processing, steering, evaluation, statistical analysis): \url{https://github.com/vkmk1/Sycophancy-Steering}. Results, figures, and analysis notebooks: \url{https://github.com/kelkarI/sycophancy-clean-results}.

\FloatBarrier
\section{Use of AI Assistants}
Claude (Anthropic) assisted with code development, statistical analysis, and manuscript drafting. Codex (OpenAI) assisted with literature checking and manuscript revision. Astra (OpenAI, via Codex) performed the retrospective, non-blinded target-fact grading described in Appendix~\ref{app:probes}. This review added annotations to existing response excerpts without changing the saved generations or primary experimental measurements. The human authors are responsible for experimental design, data interpretation, and the final scientific claims.

\end{document}